\documentclass[letterpaper, 10 pt, journal, twoside]{IEEEtran}  

\IEEEoverridecommandlockouts                              

\usepackage{times}
\usepackage{amsmath}
\usepackage{multicol}
\usepackage{amsfonts}
\usepackage{graphicx}
\usepackage{algorithm}
\usepackage{subcaption}
\usepackage{balance}
\usepackage{array}
\usepackage{makecell}
\usepackage[noend]{algpseudocode}
\DeclareMathOperator*{\argmin}{arg\,min}
\DeclareMathOperator*{\argmax}{arg\,max}

\usepackage{xcolor}
\usepackage{booktabs}

\usepackage{color}

\newcommand{\xt}{x_t}
\newcommand{\xtpo}{x_{t+1}}
\newcommand{\ut}{u_t}
\newcommand{\dut}{\delta u_t}

\newcommand{\bsig}{\bar{\sigma}^2}

\usepackage[font=small]{caption}

\usepackage{hyperref}

\title{Model-based Generalization under Parameter \\ Uncertainty using Path Integral Control}

\author{
                    Ian Abraham$^{1,2}$,
                    Ankur Handa$^1$,
                    Nathan Ratliff$^1$,
		            Kendall Lowrey$^{1,3}$,
                    Todd D. Murphey$^2$
                    and Dieter Fox$^{1,3}$%
       \thanks{
           $^1$ NVIDIA, USA
       }%
       \thanks{
           $^2$ Northwestern University, Evanston, IL 60208, USA
       }%
        \thanks{
            $^3$ University of Washington, Seattle, WA 98105, USA
        }%
        \thanks{
            {\tt\footnotesize{ email : i-abr@u.northwestern.edu, ahanda@nvidia.com, nratliff@nvidia.com, kendall.lowrey@gmail.com, t-murphey@northwestern.edu, dieterf@nvidia.com}}
        }%
}

\author{            Ian Abraham$^{1,2}$,
                    Ankur Handa$^1$,
                    Nathan Ratliff$^1$,
		            Kendall Lowrey$^{1,3}$,
                    Todd D. Murphey$^2$
                    and Dieter Fox$^{1,3}$%
\thanks{Manuscript received: September, 10, 2019; Revised November, 15, 2019;
Accepted January, 15, 2020.}
\thanks{This paper was recommended for publication by Dongheui Lee upon
evaluation of the Associate Editor and Reviewers' comments.
This work was carried out as part of an internship at NVIDIA.}
\thanks{
    $^1$ NVIDIA, USA
}%
\thanks{
    $^2$ Northwestern University, Evanston, IL 60208, USA
}%
 \thanks{
     $^3$ University of Washington, Seattle, WA 98105, USA
 }%
 \thanks{
     {\tt\footnotesize{ email : i-abr@u.northwestern.edu, ahanda@nvidia.com, nratliff@nvidia.com, kendall.lowrey@gmail.com, t-murphey@northwestern.edu, dieterf@nvidia.com}}
 }%
\thanks{Digital Object Identifier (DOI): see top of this page.}
}

\begin{document}

\refstepcounter{figure}
\makeatletter
\let\@oldmaketitle\@maketitle
\renewcommand{\@maketitle}{\@oldmaketitle
    \centerline{
        \includegraphics[width=0.8\linewidth]{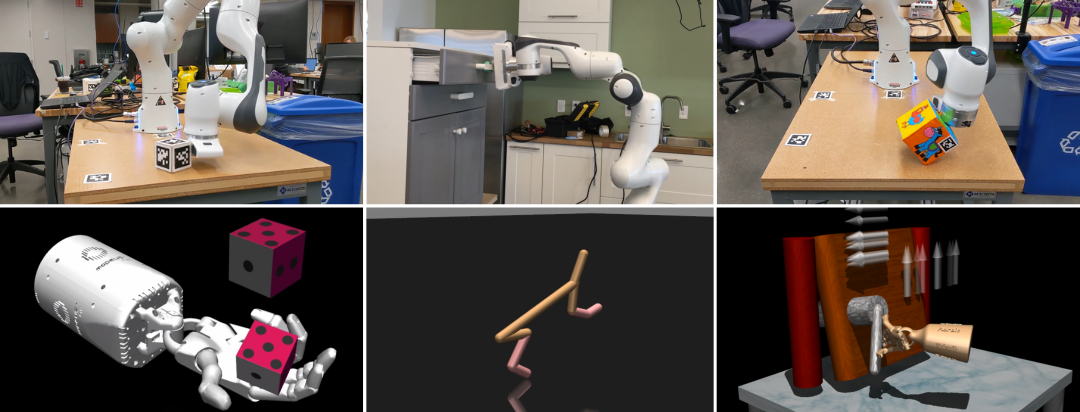}
    }
   \small  Fig.~\thefigure.
  Environments where we test our method on model-based control for generalizing under
  parameter uncertainty.  Top row illustrates the experimental test beds where we test object reconfiguration, opening drawers, and nonprehensile manipulation under parameter uncertainty. Bottom row illustrates a sample set of the simulated environments where we tested our method. From left to right we have the Shadow hand manipulating a dice, a half-cheetah robot performing a backflip, and the Adroit hand opening a door.
  \label{fig:title} \medskip \vspace{-10pt}}
\makeatother

\maketitle

\begin{abstract}

    This work addresses the problem of robot interaction in complex environments where online control and adaptation is
    necessary. By expanding the sample space in the free energy formulation of path integral control, we
    derive a natural extension to the path integral control that embeds uncertainty into action and
    provides robustness for model-based robot planning. Our algorithm is applied to a diverse set of tasks using
    different robots and validate our results in simulation and real-world experiments. We further show that our method
    is capable of running in real-time without loss of performance. Videos of the experiments as well as additional
    implementation details can be found at {\color{blue} \url{https://sites.google.com/view/emppi} }.

\end{abstract}


\section{Introduction}

    \IEEEPARstart{A}{s} the complexity of tasks and environments that robots are expected to work in increases, so will the need for
    robots to rapidly learn from experience, adapt to sensory input, and model their environments for planning. Robots
    can not solely learn from experience as this can be computationally and energetically costly, requiring many trials
    which can cause damage to the robot over time. Furthermore, experience-based learning may not generalize to
    immediate changes in the environment. Similarly, robots can not solely spend their time modeling and planning in the
    environment as the interactions can be complex and difficult to model online. This work presents a method which
    provides a solution to these issues through optimizing and adapting an ensemble of physics simulators that capture the
    underlying structure and complexities of the world while adapting to physical variations in an online model-based
    control paradigm.

    Learning robot tasks in physics simulators and applying the learned skills to the real world (also known as
    sim-to-real) has been studied quite extensively~\cite{tan18rss, peng2018sim, jakobi1995noise, tobin2017iros}. All
    the learning occurs on detailed physics simulators which handles complex friction, contact, and multi-body
    interactions. These simulators often run faster than real-time, allowing for additional computation to occur. The
    simulated robot can explore and test skills in the simulation before attempting the task in the real world, avoiding
    damage to itself and the environment. A common problem with simulated learning is poor performance of learned skills
    when transferred into the real world. The poor performance is attributed to what is known as a \emph{reality
    gap}~\cite{jakobi1995noise} where imprecise physical parameters and physics interactions in the simulated world
    render simulated learned skills useless. Existing work tries to overcome these faults by diversifying the physical
    parameters of the world~\cite{peng2018sim}, improving the accuracy of the physics simulators~\cite{tan18rss}, or
    updating the simulator parameters after real-world experience~\cite{chebotar2018closing}. While currently the state
    of the art, these methods work in two-stages: train in simulation, then apply to real-world. Our work uses the fact
    that current simulators can perform faster than real time, further improving the capabilities of sim-to-real in an
    online setting where we formulate stochastic model-based control with an ensemble of physics simulators
    with parameter variations, enabling us to synthesis a control signal for robotic systems that naturally encodes
    parameter uncertainty and adapting the simulator parameters online.

    Our approach expands upon the free energy formulation of model-predictive path integral control
    (MPPI)~\cite{williams2017information, williams2016aggressive, williams2018robust, theodorou2010generalized} by
    naturally encoding variations in physical parameters and structure in physics engine into online control synthesis.
    Any synthesized control signal is able to generalize to variations in the simulated worlds while adapting to the
    uncertainty to solve the task. The control synthesis generates actions that initially are conservative, eventually
    adapting and becoming more exploitative as sensory input is acquired. Thus, our contribution is an online ensemble
    model-based control algorithm that completes tasks while being robust and adaptive to parameter uncertainty.
    Simulated and real-world examples validate our approach for generalizing to model-based uncertainty.

    The paper is structured as follows: Section~\ref{sec:related-work} provides a discussion on related work,
    Section~\ref{sec:problem-statement} formalizes and introduces the problem statement,
    Section~\ref{sec:mppi} derives the algorithm used in this paper,
    results and conclusion are presented in Section~\ref{sec:results} and~\ref{sec:conclusion}.

\section{Related Work} \label{sec:related-work}

    Current state-of-the-art in simulated robot skill learning works in two stages: first learn in simulation and then
    apply the learned skills in the real-world~\cite{peng2018sim, chebotar2018closing, tobin2017iros}. These methods
    work by using reinforcement learning techniques~\cite{schulman2015trust, schulman2017proximal} in simulation in
    order to synthesize a policy for a task. These methods often fail due to a mismatch between physical parameters in
    the real and simulated world. The work in~\cite{chebotar2018closing} tries to overcome the mismatch issue by
    ``closing the loop'' on the learning process. This loop-closure effectively allows the simulation to be updated at
    every attempt at a task in the real-world. Other work presents an approach for the model-mismatch problem using
    robust control and fine-tuning of a simulation to better predict physical phenomena~\cite{tan18rss} or
    learning policies from an ensemble of simulated environments~\cite{rajeswaran2017ICLR}. Our work
    combines these two stages of learning in simulation and adapting based on real-world
    experience. As a result, we are able to solve tasks immediately by synthesizing controls that generalize to
    various simulated environmental parameters.

    Our approach mirrors model-based optimal control method (in particular ensemble model-based
    control~\cite{tassa2014control,li2004iterative, tassa2012synthesis, stulp2011learning}) which have existed for some
    time. Specifically, we use an ensemble~\cite{becker2010motion, mordatch2015ensemble} of physics simulators in
    receding horizon to generate an optimal control sequence which we apply to the robot.  Prior work uses motion
    primitives and policies~\cite{becker2010motion, stulp2011learning}, or standard trajectory optimization with PD
    control feedback to handle the ensemble models~\cite{mordatch2015ensemble}. The main difference is that we utilize
    the free energy formulation for a stochastic control problem to directly incorporate the uncertainty in the ensemble
    into synthesis of the control. This enables us to handle the uncertainty as a natural extension to stochastic
    model-based control. Furthermore, utilizing stochastic control allows us to handle discontinuous events without
    concern which is often difficult to handle for common model-based
    controllers~\cite{tassa2014control,li2004iterative,todorov2005generalized}. Last, our approach is able to adapt to
    real-world experience through online adaptation of the parameters and receding horizon control which allows for
    reactive planning while generalizing to local variations in the physical parameters of the simulated world.

    The work in~\cite{Yu-RSS-17, yu2018policy} first introduced the idea of using simulators with system identification
    to improve the performance of sim-to-real transfer. The mention work shows that the reality gap can be overcome when
    a policy is trained to be invariant to the variations found in the physical parameters. The main difference between
    this work and our own is that we apply this approach online and illustrate that complex systems can be modeled and
    controlled in real-time in a receding horizon controller formulation that can generalize to the variations in the
    physics simulators. In contrast,~\cite{Yu-RSS-17} uses a series of learned policies using reinforcement learning and
    an online system identification function to counteract the reality gap. Similarly,~\cite{yu2018policy} learns a
    family of policies and then searches for the best policy. These methods require the need to learn the policies
    a-priori from data which is a rigid process and difficult to adapt online. We extend these ideas to online
    model-based receding horizon control where we weigh in the variations in the simulated physical worlds and update
    them online to provide informed and reactive control.

    \begin{figure*}[ht!]
        \centering
        \centerline{
            \includegraphics[width=0.2\linewidth]{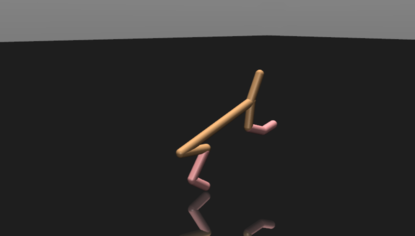}
            \includegraphics[width=0.2\linewidth]{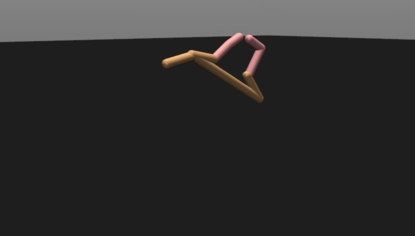}
            \includegraphics[width=0.2\linewidth]{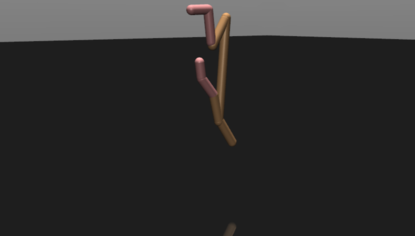}
            \includegraphics[width=0.2\linewidth]{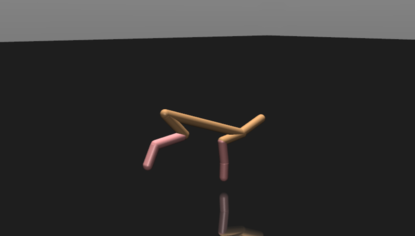}
        } \vspace{1mm}
        \centerline{
            \includegraphics[width=0.2\linewidth]{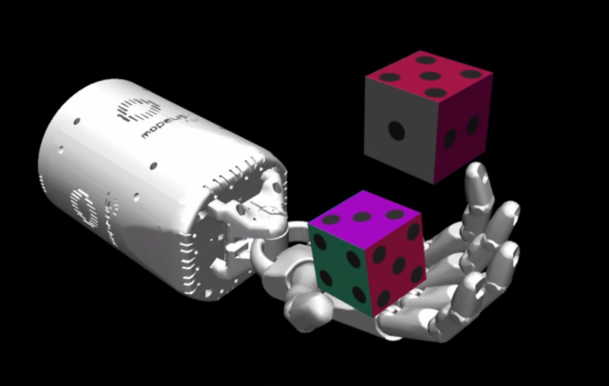}
            \includegraphics[width=0.2\linewidth]{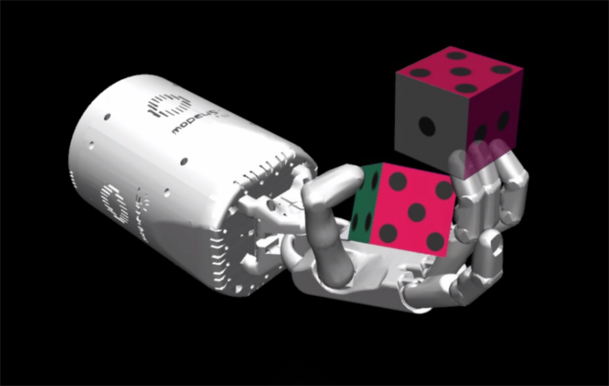}
            \includegraphics[width=0.2\linewidth]{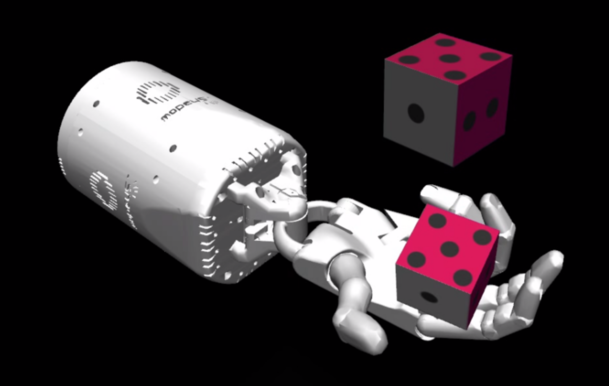}
            \includegraphics[width=0.2\linewidth]{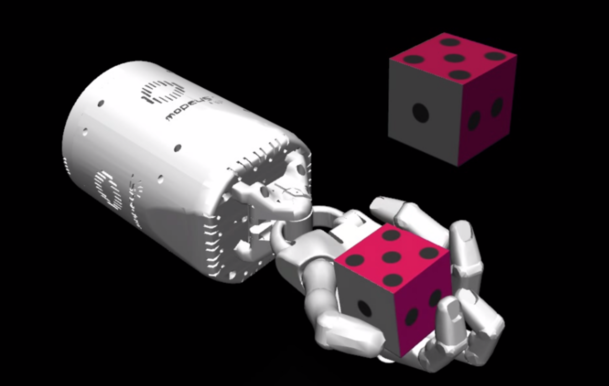}
        } \vspace{1mm}
        \centerline{
            \includegraphics[width=0.2\linewidth]{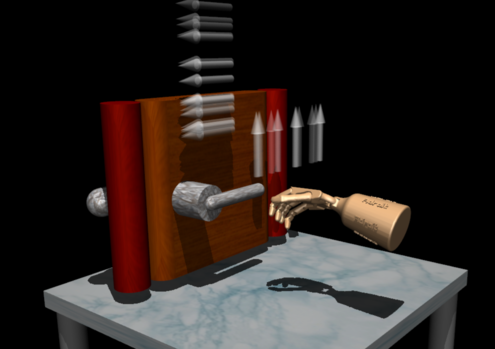}
            \includegraphics[width=0.2\linewidth]{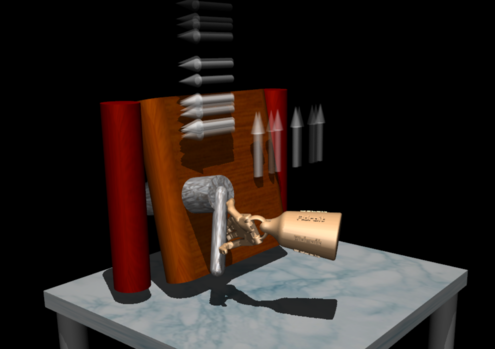}
            \includegraphics[width=0.2\linewidth]{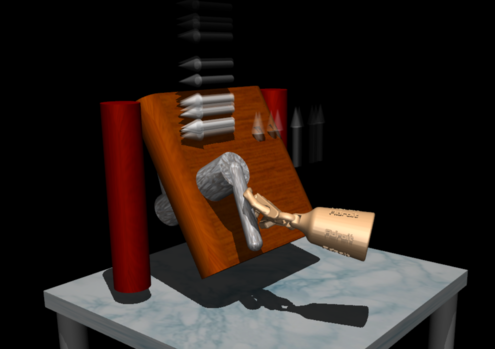}
            \includegraphics[width=0.2\linewidth]{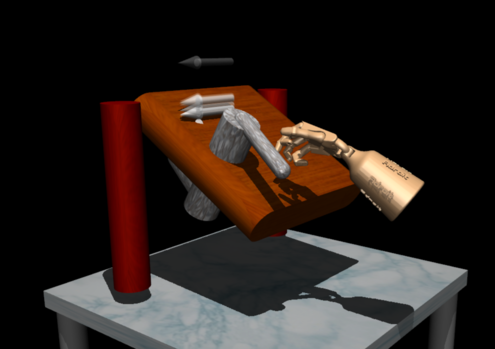}
        } \vspace{1mm}
        \caption{
		      Time series images of (top) half-cheetah backflip where the link masses and joint damping values are unknown,
              (middle) inhand manipulation of the dice while the mass of the dice is unknown, and (bottom)
              opening a door where the articulation of the door is unknown.
              Using our method, all tasks are successful regardless of uncertainty in the parameters of the simulated environments.
          }
        \label{fig:sim_results}
	\end{figure*}

\section{Problem Statement} \label{sec:problem-statement}

    Let us first consider the partially unknown dynamics \footnote{We assume we at least know the type of robot and how
    it is articulated.} of a robot and the objects in the environment as a stochastic dynamical system of the form

    \begin{equation}
        \xtpo = f(\xt, \ut + \dut ; \theta)
    \end{equation}
    where $\xt \in \mathbb{R}^n$ is the state of the system at time $t$, $\ut \in \mathbb{R}^m$ is the control input to
    the system, $\dut \sim \mathcal{N}(0, \Sigma)$ is normally distributed noise with variance $\Sigma \in \mathbb{R}^{m
    \times m}$, $f : \mathbb{R}^n \times \mathbb{R}^m \times \mathbb{R}^p \to \mathbb{R}^n$ is the transition model, and
    $\theta \sim p(\bar{\theta})$ is a set of physical parameters sampled from a distribution $p(\bar{\theta})$ with
    mean $\bar{\theta} \in \mathbb{R}^p$. Let us assume that we can sample a set of $N$ parameters $\{ \theta_n
    \}_{n=1}^{N}$ such that each parameter describes a candidate transition model. The problem is to find the sequence
    of controls $u_t \forall t \in \{0, \ldots, T-1 \}$ for a robotic system such that it solves a task specified by the
    cost function

    \begin{equation}
        \mathcal{S}(x_0, x_1, \ldots, x_T) = \mathcal{S}(v) =  m(x_T) + \sum_{t=0}^T \ell (x_t)
    \end{equation}
    subject to uncertainty in the transition model parameters $\theta$. Here, $\ell(x) : \mathbb{R}^n \to \mathbb{R}$
    and $m(x) : \mathbb{R}^b \to \mathbb{R}$ are the state dependent running cost and terminal cost respectively for the
    value function $\mathcal{S}(v)$, and $v = \{ v_0, v_1, \ldots, v_{T-1} \}$ is the sequence of stochastic controls
    $v_t = u_t + \delta u_t$ that generate the states $x_t$. The specified problem is equivalent to the stochastic
    optimal control problem

    \begin{align}\label{eq:control_problem}
        u^\star_t & =  \argmin_{u_t \forall t \in [0, T-1]} \mathbb{E}_{ \theta \sim p(\bar{\theta}), \mathbb{Q}}\left[ \mathcal{S}(v) \right] \\
        & = \argmin_{u_t \forall t \in [0, T-1]} \mathbb{E}_{ \theta \sim p(\bar{\theta}), \mathbb{Q}}\left[ m(x_T) + \sum_{t=0}^{T-1} \ell(x_t ) \right] \nonumber
    \end{align}
    where $\mathbb{E}$ is the expectation operator with respect to the distribution $p(\bar{\theta})$ and
    the open-loop control distribution $\mathbb{Q}$, i.e., the distribution of admissible control sequences.

    Assuming that the true physical system parameters reside within a local minima\footnote{Local minima refers to the,
    often common, coupling between physical parameters that result in similar or apparent behavior of rigid bodies given
    another set of distinct parameters.} of the set of parameters $\{ \theta_n \}$, a solution to
    (\ref{eq:control_problem}) will be a sequence of controls that can generalize amongst the variations in the physical
    parameters. As an aside, we treat each of these parameters as a particle (similar to a particle filter) that
    describes the simulated world. Then $p(\theta_n)$\footnote{This expression is used as short hand for the likelihood
    $p(\theta_n \mid \bar{\theta})$.} is viewed as the weight of the particle and how likely the particle is in
    representing the real world. The optimal sequence of controls would then be a sample from an optimal control
    distribution $\mathbb{Q}^\star$ which is used in the following Section~\ref{sec:mppi} in developing the proposed
    algorithm.

    \setlength\tabcolsep{2pt} 
    \begin{table*}[]
        \centering
        \begin{tabular}{ccccccc}
        \toprule
        Task name & Dimensions & \makecell{Running cost $\ell(x_t)$\\ Terminal Cost $m(x_T)$} & Uncertain Param.
                                                & Init. Distr. & Samples & Hyper Param.\\ \midrule
        \makecell{Half-cheetah\\ backflip} &
        \makecell{$x \in \mathbb{R}^{18}$ \\ $u \in \mathbb{R}^6$} & $10 x_\text{vel}^2 + 60 (x_\text{rot} + 2 \pi)^2$  & \makecell{body mass x6,\\ joint damping x7} & \makecell{uniform$(1,6)$\\uniform$(1,6)$}
                                & \makecell{$N=20$\\$K=4$}
                                & \makecell{$\Sigma= 0.5 \mathbf{I}$ \\ $\lambda = 0.1$ \\ $T=40$ \\ $\bsig=0.01$}\\ \hline
        \makecell{Inhand \\ Manipulation \\ (Shadow)} &
        \makecell{$x \in \mathbb{R}^{61}$ \\ $u \in \mathbb{R}^{20}$} & \makecell{
                                        $\! \begin{aligned} & 400 \Vert p_\text{palm}- p_\text{obj}\Vert_2  \\
                                        &+ 10 \Vert \log(R_\text{obj}^{-1}R_\text{targ}) \Vert_\text{Fr} + 0.1 \Vert x_\text{vel} \Vert_2 \end{aligned}$
                                        \\ \hline\hline
                                        $m(x_T) = 100 \Vert p_\text{palm} - p_\text{obj} \Vert_2$
                                        }  & \makecell{dice mass, \\ finger actuator \\ gains x18}
                                        & \makecell{uniform$(0.01, 0.1)$\\uniform$(0.01,5)$}
                                        & \makecell{$N=20$ \\ $K=2$}
                                        & \makecell{$\Sigma= 0.8 \mathbf{I}$ \\ $\lambda = 0.001$ \\ $T=20$ \\ $\bsig=1$} \\ \hline
        \makecell{Opening \\Door (Adroit)} &
        \makecell{$x \in \mathbb{R}^{60}$ \\ $u \in \mathbb{R}^{28}$}
                                        &  $80 \Vert p_\text{palm} - p_\text{handle} \Vert_2
                                        + 0.01 \Vert x_\text{vel} \Vert_2
                                        + 100 x_\text{hinge} $  & \makecell{door hinge axis \\and position}
                                        & \makecell{binomial$(1,0.5)$ \\ uniform$(0.1, 0.5)$}
                                        & \makecell{$N=40$\\$K=1$}
                                        & \makecell{$\Sigma= 0.2 \mathbf{I}$ \\ $\lambda = 0.8$ \\ $T = 30$ \\ $\bsig=0.01$} \\ \hline
        \makecell{Opening \\ Cabinet \\ (Franka)} &
        \makecell{$x \in \mathbb{R}^{20}$ \\ $u \in \mathbb{R}^7$}
                                        & \makecell{
                                        Reach: $\! \begin{aligned}
                                        & 200 \Vert \log(R_\text{gripper}^{-1}R_\text{handle}) \Vert_\text{Fr} \\
                                        & \qquad + 200 \Vert p_\text{gripper} - p_\text{handle}\Vert
                                        + 0.01 \Vert x_\text{vel} \Vert_2 \end{aligned}$\\ \hline\hline
                                        Pull: $\! \begin{aligned} & 100 p_\text{x, handle}
                                        + 40 \Vert p_\text{gripper} - p_\text{handle}\Vert \\
                                        & \qquad + 40 \Vert \log(R_\text{gripper}^{-1}R_\text{handle}) \Vert_\text{Fr}
                                        \end{aligned}$}
                                        & \makecell{drawer hinge axis\\  and position}
                                        & \makecell{normal$([1,0,0], 0.6 \mathbf{I})$ \\ uniform$(-0.2,0.2)$}
                                        & \makecell{$N=10$ \\$K=2$}
                                        & \makecell{$\Sigma= 1.0 \mathbf{I}$ \\ $\lambda = 0.8$ \\ $T=20$ \\ $\bsig=0.01$} \\ \hline
        \makecell{Nonprehensile \\ Manipulation \\ (Franka)} &
        \makecell{$x \in \mathbb{R}^{20}$ \\ $u \in \mathbb{R}^7$}
                                        & \makecell{
                                        $\Vert p_\text{gripper} - p_\text{object}\Vert +
                                        2 \Vert p_\text{target} -p_\text{object} \Vert
                                        + \sum F_\text{contact}$}
                                        & \makecell{object mass \\  and friction coeff.}
                                        & \makecell{normal$(0.25, 0.1)$ \\ normal$(7.5\mathrm{e}{-3}, 0.01)$}
                                        & \makecell{$N=10$ \\$K=1$}
                                        & \makecell{$\Sigma= 0.08 \mathbf{I}$ \\ $\lambda = 0.02$ \\ $T=8$ \\ $\bsig=0.2$}\\ \hline
        \makecell{Object \\ Reconfiguration \\ (Franka)} &
        \makecell{$x \in \mathbb{R}^{20}$ \\ $u \in \mathbb{R}^7$}
                                        & \makecell{
                                        $\Vert p_\text{gripper} - p_\text{object}\Vert + 10 \Vert q_\text{target} -q_\text{object} \Vert + \sum F_\text{contact}$}
                                        & \makecell{object mass \\  and friction coeff.}
                                        & \makecell{normal$(1.0, 0.1)$ \\ normal$(7.5\mathrm{e}{-3}, 0.01)$}
                                        & \makecell{$N=10$ \\$K=1$}
                                        & \makecell{$\Sigma= 0.08 \mathbf{I}$ \\ $\lambda = 0.02$ \\ $T=8$ \\ $\bsig=0.2$}\\ \hline
        \bottomrule
        \end{tabular}
        \caption{
            Task specifications for each example presented in this work.
            The running cost for each task is shown (terminal cost only shown if used).
            The uncertain parameter in each task is described whose initial sampling distribution is
            described in the following column respectively.
            The variables $p,q,F,R$ refer to the position, quaternion, force, and rotation matrix.
            The parameter $\bsig$ refers to the diagonal variance used in the likelihood model (empirically chosen).
            Last, the number of simulator samples $N$ and trajectory samples per parameter $K$ is shown for each task.
        }
        \vspace{-5mm}
        \label{table:param}
    \end{table*}

\section{Ensemble Model-Predictive Path Integral Control (EMPPI)} \label{sec:mppi}

    In this section, we present ensemble model-predictive path integral control (EMPPI) as a natural extension to the
    information theoretic approach to the path integral control problem~\cite{williams2017information}. The concept
    behind path integrals is to indirectly solve for (\ref{eq:control_problem}) by minimizing the free energy of the
    control system~\cite{theodorou2012relative}. For a full derivation of model predictive path integral
    control, we refer the reader to \cite{williams2017information}. We start with the definition of the free energy
    formulation that is introduced in \cite{williams2017information}. Here, the sample space of the free energy
    formulation of the control system is expanded in order to incorporate model uncertainty into the control problem.
    This enables us to derive the controller that best generalizes to the parameter uncertainty as well as synthesize
    actions that encourage robustness given the uncertainty. Using Jensen's inequality and importance sampling, the free
    energy\footnote{We refer the reader to the work in~\cite{williams2017information, theodorou2012relative} for more
    information on path integral control and free energy formulations.} of the stochastic control system in question is
    defined as
    \begin{align}\label{eq:free_energy}
        \mathcal{F}(v) &= - \lambda \log \left( \mathbb{E}_{p(\bar{\theta}), \mathbb{P}} \left[ \exp \left( -\frac{1}{\lambda} \mathcal{S}(v)\right)\right] \right) \nonumber \\
         & \le - \lambda \mathbb{E}_{p(\bar{\theta}), \mathbb{Q}} \left[ \log \left( \frac{p(v)}{q(v)} \exp \left( -\frac{1}{\lambda} \mathcal{S}(v)\right) \right) \right] \nonumber \\
         & \le \mathbb{E}_{p(\bar{\theta})} \left[ \mathbb{E}_\mathbb{Q}\left[ \mathcal{S}(v) \right]
                    - \lambda \mathbb{E}_\mathbb{Q}\left[ \log \left( \frac{p(v)}{q(v)}\right)\right] \right] \nonumber \\
         & \le \mathbb{E}_{p(\bar{\theta})} \left[ \mathbb{E}_\mathbb{Q}\left[ \mathcal{S}(v) \right] + \lambda D_\text{KL}\left(\mathbb{Q} \Vert \mathbb{P} \right)\right]
    \end{align}
    where $\mathbb{P}$ and $\mathbb{Q}$ are the control noise of the uncontrolled system distribution and
    the open loop control distribution defined by the probability density functions
    \begin{align} \label{eq:pdfs}
        p(v) &= \prod_{t=0}^{T-1} ((2 \pi)^m \vert \Sigma \vert)^{-\frac{1}{2}}
            \exp\left( -\frac{1}{2}v_t^\top\Sigma^{-1} v_t \right) \text{ and}   \\
        q(v) &= \prod_{t=0}^{T-1} ((2 \pi)^m \vert \Sigma \vert)^{-\frac{1}{2}}
            \exp\left( -\frac{1}{2}(v_t-u_t)^\top\Sigma^{-1} (v_t -u_t) \right) \nonumber
    \end{align}
    respectively.
    Here, $\Sigma \in \mathbb{R}^{m \times m}$ is the covariance of the control signal, $\lambda$ is a positive scalar
    variable known as the temperature~\cite{williams2017information, chebotar2017path}, and $D_\text{KL}(\mathbb{Q}
    \Vert \mathbb{P})$ is the Kullback-Leibler (KL) divergence measure between $\mathbb{Q}$ and $\mathbb{P}$. From
    (\ref{eq:free_energy}), it can be seen that choosing $\frac{p(v)}{q(v)} \propto 1/\exp\left( -\frac{1}{\lambda}
    \mathcal{S}(v)\right)$ reduces the KL-divergence to $-\mathbb{E}[\mathcal{S}(v)]$ which converts the free energy
    into a constant and an equality which corresponds to the control objective in (\ref{eq:control_problem}). This
    suggests we can solve the stochastic optimal control problem by instead solving for the corresponding sample-based
    optimization:
    \begin{equation} \label{eq:surrogate_optimization}
        u^\star_t = \argmin_{u_t} \mathbb{E}_{p(\bar{\theta})}
                    \left[ D_\text{KL}\left( \mathbb{Q}^\star \Vert \mathbb{Q}\right) \right]
    \end{equation}
    which is equivalent to solving for the control problems specified in (\ref{eq:control_problem}) and
    (\ref{eq:free_energy}) where $\mathbb{Q}$ is replaced with the optimal distribution $\mathbb{Q}^\star$ and the
    uncontrolled system noise distribution $\mathbb{P}$ becomes the open-loop control distribution $\mathbb{Q}$.

    Since we do not have access to $\mathbb{Q}^\star$ and thus can not directly sample from the optimal distribution,
    $\mathbb{Q}^\star$ is defined through its probability density function
    \begin{equation}
        q^\star(v) = \frac{1}{\eta} \exp\left( -\frac{1}{\lambda} \mathcal{S}(v) \right) p(v).
    \end{equation}
    Using importance sampling by multiplying by $p(s)/p(s)$, we can decoupling the expectation over the parameter
    distribution and separate (\ref{eq:surrogate_optimization}) into two parts. \footnote{This is possible as the
    optimization is not over the parameters $\theta$.} The resulting optimization in (\ref{eq:surrogate_optimization})
    using the density functions in (\ref{eq:pdfs}) is defined as the closed form solution
    \begin{align}\label{eq:closed_form}
        u^\star_t & = \mathbb{E}_{p(\bar{\theta})} \left[ \argmin_{u_t} \int_\Omega q^\star(v) \log\left( \frac{q^\star(v)}{p(v)} \frac{p(v)}{q(v)} \right) dv \right] \nonumber \\
        & = \mathbb{E}_{p(\bar{\theta})} \left[ \argmax_{u_t} \int_\Omega q^\star(v) \log\left( \frac{q(v)}{p(v)}\right) dv\right] \nonumber \\
        & = \mathbb{E}_{p(\bar{\theta})} \left[ \int_\Omega q^\star(v) v_t dv\right]
    \end{align}
    where going from the first step to the second step in (\ref{eq:closed_form}) is a result of separating the inner
    integral into two parts where one of the parts does not depend on $u_t$, and $\Omega$ is the sample space of
    $\mathbb{R}^m \times T-1$. Using iterated expectations and the change of variable $v_t = u_t + \delta
    u_t$, we write Eq.(\ref{eq:closed_form}) as
    \begin{equation*}
        u^\star_t = \sum_{n=1}^N
        \mathbb{E}_{p(v)}\left[ \frac{1}{\eta} \exp \left( -\frac{1}{\lambda} S(v) \right)v_t \ \Big\vert \ \theta_n \right] p(\theta_n)
    \end{equation*}
    where the inner expectation is conditioned on a single parameter $\theta_n$ through the
    forward prediction of the physics model.
    Generating samples from $p(v)$, we obtain the recursive solution
    \begin{equation}
        u^\star_t \gets u_t + \sum_{n=1}^N \sum_{k=1}^K \omega(v_{t,k,n}) \delta u_{t,k,n}
    \end{equation}
    where
    \begin{align*}
        \omega(v_{t,k,n}) &=
        \frac{\exp \left(-\frac{1}{\lambda}\left(\tilde{S}(v_{t,k,n})-\beta\right) \right) p(\theta_n)}
        {\sum_{n=1}^{N}\sum_{k=1}^{K}\exp \left(-\frac{1}{\lambda}\left(\tilde{S}(v_{t,k,n})-\beta\right) \right)p(\theta_n)}, \nonumber \\
        \beta &= \argmin_{k,n} \tilde{S}(\tau_{t,k,n}) \nonumber
    \end{align*}
    is the sample weights defined as a softmax function for numerical stability~\cite{williams2017information}
    (normalizes the weight contribution from the cost function and $p(\theta)$), and
    \begin{multline*}
        \tilde{S}(v_{t, k, n}) = \sum_{t}^{T-1}\ell(x_{t,k,n})
        + \lambda u_t^\top \Sigma^{-1} \delta u_{t,k,n}
        + m(x_{T,k,n}),
    \end{multline*}
    is the value function starting at any time $t \in \left[0, T-1\right]$ of the planning horizon. Here, the random
    control $v_{t,k,n}$ is indexed based on the time, samples for the parameters, and the trajectory samples. Note that
    each trajectory sample has a constant parameter $\theta_n$ which is used to simulate forward given the control
    $v_t$. Thus the solution $u^\star_t$ is the solution that generalizes best to the set of parameters given the
    likelihood that the parameter represents the true system dynamics.

    By recursively updating the distribution $p(\theta_n)$, the control solution is able to have immediate impact on the
    performance of the task during execution. Each parameter is updated based on how well it is able to predict the
    subsequent state of the robotic system given the applied control, that is
    \begin{equation}
        p(\theta_n) \gets p(x_{t+1} \vert x_t, u_t, \theta_n) p(\theta_n) /\eta
    \end{equation}
    where $\eta$ is a normalization constant, and $p(x_{t+1} \vert x_t, u_t, \theta_n)$ is assumed a Gaussian likelihood
    function~\cite{deisenroth2011pilco} with diagonal variance parameters set in Table~\ref{table:param} . Each
    parameter is resampled with the current mean when $1/\sum_n p(\theta_n)^2 < N_\text{eff}$ where $N_\text{eff} = 0.5
    N K$ to avoid a degenerate set of weights. This effective sample size cutoff is often used with
    particle filters ~\cite{speekenbrink2016tutorial} which provides an estimate on how much each sampled parameter has
    an effect on the mean of the distribution $p(\bar{\theta})$. When this condition is met, each $\theta_n$ is
    resampled with added noise to encourage diversity in synthesis of the controller. Algorithm~\ref{alg:1} illustrates
    the pseudocode for our method.

    \begin{algorithm}
    \caption{Ensemble Model-Predictive Path Integral Control (EMPPI)}\label{alg:1}
        \begin{algorithmic}[1]
            \State \textbf{initialize:} sim $f(x, u; \theta)$, parameter distribution $p(\bar{\theta})$ with $N$ parameter samples,
            $K$ number of trajectory samples, time horizon $T$, objective $\ell$, terminal cost $m$,
            control noise $\Sigma$, temperature parameter $\lambda$, $u_t = 0 \ \forall t \ \in [0, T-1]$, real world time $\tau=0$
            \While{task not done}
                \State sample state $\bar{x}_\tau$ from robot environment
                \State set each sim $f$ with sampled state $x_{0, k, n} \gets \bar{x}_\tau$
                \For{$\theta_n \in \{\theta_1, \ldots, \theta_N \}$}
                    \For{$k \in \{1, \ldots, K \}$}
                        \For{$t \in \{0, \ldots, T-1 \}$}
                            \State $\triangleright$ sample $\delta u_{t,k,n} \sim \mathcal{N}(0, \Sigma)$
                            \State $\triangleright$ set local cost variable $s^{k,n}_t$ using $x_t, u_t$
                            \State $s^{k,n}_t = \ell(x_{t,k,n}) + \lambda u_t^\top \Sigma^{-1} \delta u_{t,k,n}$
                            \State $\triangleright$ update state with transition model
                            \State $x_{t+1,k,n} = f(x_{t,k,n}, u_t + \delta u_{t,k,n}; \theta_n)$
                        \EndFor
                        \State $\triangleright$ add terminal cost
                        \State $s^{k,n}_{T} = m(x_{T,k,n})$
                    \EndFor
                \EndFor
                \For{$t \in \{0, \ldots, T-1 \}$}
                    \State $\Tilde{S}(\tau_{t,k,n}) \gets \sum_{j=t}^{T} s^{k,n}_j$
                    \State $u_t \gets u_t + \sum_{n=1}^{N} \sum_{k=1}^{K} \omega(v_{t,k,n}) \delta u_{t, k, n}$
                \EndFor
                \State $\triangleright$ update parameter distribution from past state $x_{-1}$ and control $u_{-1}$
                \State $p(\theta_n) \gets p(x_0 \mid x_{-1}, u_{-1}, \theta_n)p(\theta_n)/ \eta$
                \State resample $p(\theta)$ if $1/\sum_n p(\theta_n)^2 < N_\text{eff}$
                \State \textbf{optional} apply filter to $u_t$
                \State apply $u_0$ to robot and shift control sequence
            \EndWhile
        \end{algorithmic}
    \end{algorithm}

\section{Results} \label{sec:results}

    In this section, we evaluate our method against a diverse set of robotic systems and tasks (both in simulation and
    in the experiments). The focus of this section is to show that our method is robust to various kinds of uncertainty
    while enabling the robotic system to complete the task at hand. Furthermore, we aim to show that our method's
    performance does not degrade when compared to model-based control with perfect information about the physical
    environment. For each simulated system we artificially added zero-mean noise ($\sigma^2 = 0.001$) to
    each state measurement to simulate real-world noisy measurements. Each example is able to successfully complete the
    task within the first initial run of the algorithm. Subsequent runs are done to analyze the convergence of the
    parameter distribution. We refer the reader to the paper site {\color{blue}
    \url{https://sites.google.com/view/emppi} } for an ablation study on the parameters $N$ and $K$ of EMPPI using the
    cart pole swingup task where the mass and inertia of the pole are uncertain as well as a comparison of EMPPI to MPPI
    with a learned model.

    \subsection{Simulated Results}

        \textbf{Half-cheetah Backflip:} We evaluate EMPPI on a variety of simulated systems (see
        Fig.~\ref{fig:sim_results}). The first example is using the half-cheetah robot. It is assumed that we do not
        have full knowledge of the half-cheetah's joint damping as well as body masses (see Table~\ref{table:param} for
        parameter detail). The goal is for the half-cheetah to attempt a full backflip while there is uncertainty in the
        parameters of the robot itself. The state space is the joint poses and velocities as well as the $x-y$
        position and rotation of the largest body. The control space is the applied joint torques to the half-cheetah.

        \begin{figure}
            \centering
            \includegraphics[width=\linewidth]{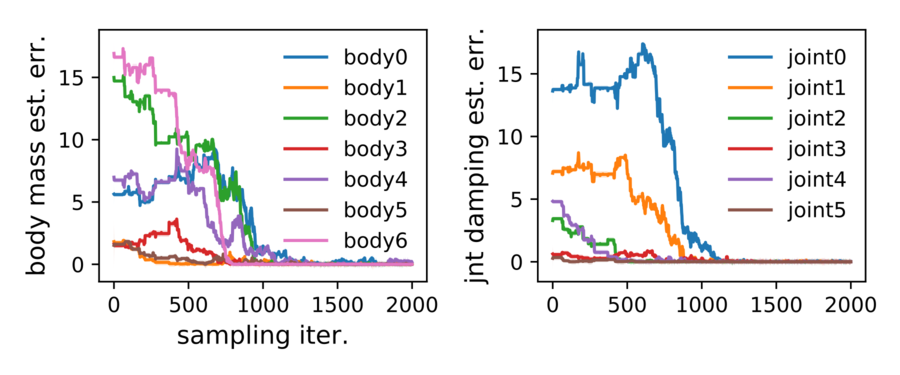}
            \caption{
                    Online improvement of half-cheetah body masses (kg) and joint damping (kg/s) values during backflip maneuver.
                    Each line corresponds to the squared error of the individual estimate of the half-cheetah's joint
                    damping values and body masses.
            }
            \label{fig:half_cheetah_results}
        \end{figure}

        Figure~\ref{fig:half_cheetah_results} shows that our method is able to predict the true parameters of the
        half-cheetah's body mass and joint damping values within $1000$ time steps. The robot is run until
        it can accomplish a backflip, the system is reset and continues trying to backflip. Initially, the proposed
        method generates a control signal that attempts to backflip. During this process, state data is collected which
        updates the parameter belief. Note that the half-cheetah is able to successfully backflip within $400$ time
        steps which suggests only a partial estimate of the true parameters is needed for successful backflip. We reset
        the robot to the start position once the backflip is accomplished. Therefore, EMPPI is able to
        accomplish the task regardless of the imperfect estimates of each link mass and joint damping. This is a direct
        result of formulating the control problem with the uncertainty in the physics simulation parameters. With this
        formulation, each synthesized action is robust to the uncertainty of the physical parameters while exploiting
        the dynamic structure provided by the physics engine.

        \begin{figure}
            \centering
            \includegraphics[width=0.8\linewidth]{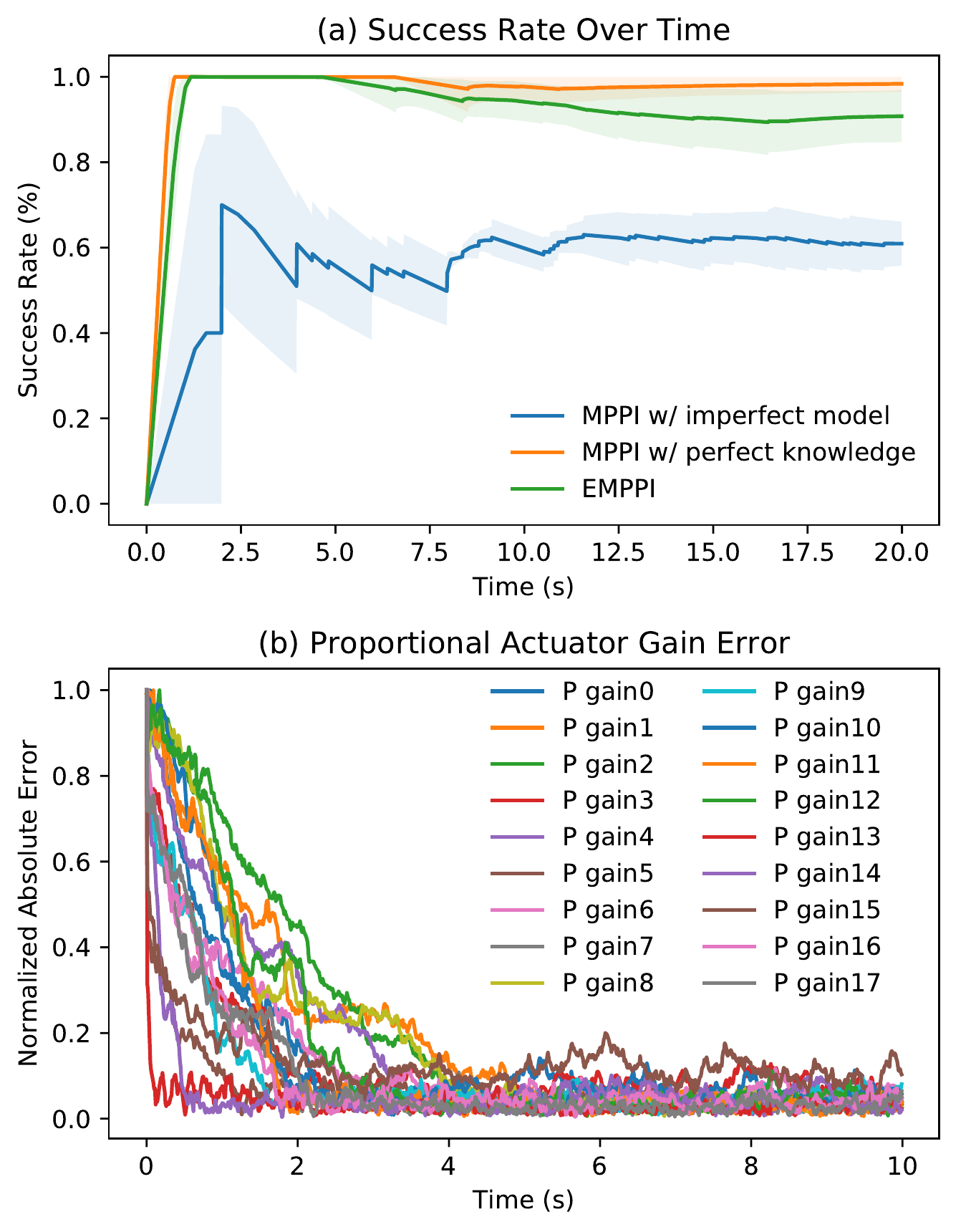}
            \caption{
                Comparison of our method (EMPPI) on the Shadow hand dice manipulation task when there is uncertainty in
                the proportional actuator gains of the finger of the Shadow hand. Success rate is
                calculated as the sum of total successful attempts at manipulating the dice divided by the total
                attempts made within the allotted time (variance derived from testing over 5 trials). (a) Our method is
                shown to maintain similar success rates compared to MPPI with perfect information. Initializing MPPI
                with a model with incorrect parameters results in significant performance loss. (b) EMPPI is able
                to achieve accurate estimate the uncertain parameters within $4$ seconds.
            }
            \vspace{-5mm}
            \label{fig:act_gain_results}
        \end{figure}

        \begin{figure*}[ht!]
            \centerline{
                \includegraphics[width=0.22\linewidth]{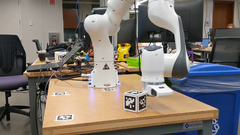}
                \includegraphics[width=0.22\linewidth]{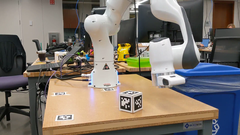}
                \includegraphics[width=0.22\linewidth]{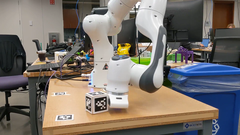}
                \includegraphics[width=0.22\linewidth]{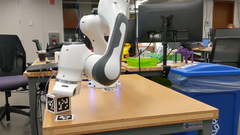}
                } \vspace{1mm}
            \centerline{
                \includegraphics[width=0.22\linewidth]{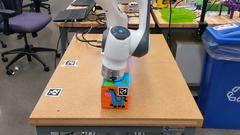}
                \includegraphics[width=0.22\linewidth]{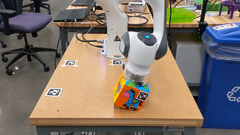}
                \includegraphics[width=0.22\linewidth]{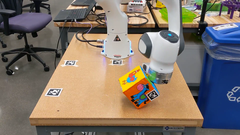}
                \includegraphics[width=0.22\linewidth]{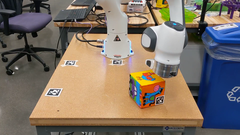}
                } \vspace{1mm}
            \centerline{
                \includegraphics[width=0.22\linewidth]{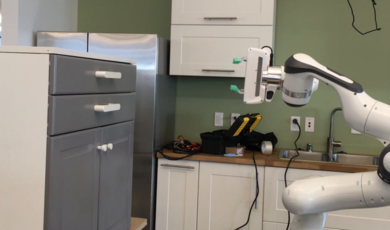}
                \includegraphics[width=0.22\linewidth]{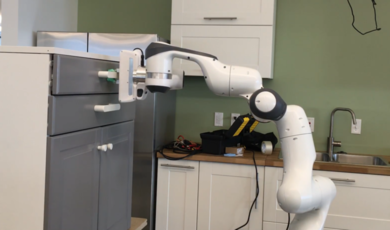}
                \includegraphics[width=0.22\linewidth]{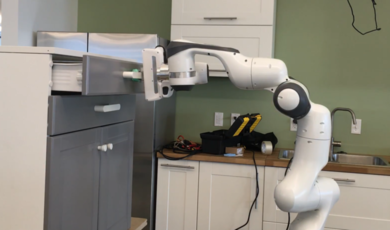}
                \includegraphics[width=0.22\linewidth]{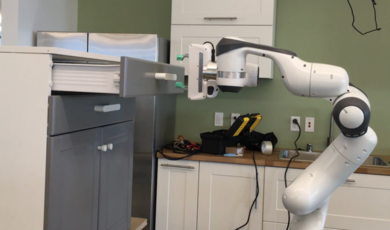}
                }
            \caption{
                Time-series images of real-world experiments of EMPPI using Franka Panda robot. Starting at the top, we
                show the Panda arm pushing a block whose mass and sliding friction is unknown towards the bottom of the
                target tag at the lower left of the image. Next is the Panda arm rotating a block whose mass and
                friction is unknown towards a desired target configuration. Last, is the Panda arm opening a drawer
                where the articulation of the drawer is unknown. Each of the examples are done in real-time on a
                single roll-out of robot running EMPPI.
            }
            \label{fig:franka_cabinet}
        \end{figure*}

        \textbf{Shadow Hand Manipulation for EMPPI Comparison: } The following example uses the Shadow Dexterous robot
        hand for inhand manipulation of a dice~\cite{plappert2018multi}. The goal is for the robot to move
        the dice to the target orientation ($<0.15$ $\ell_2$ error in quaternions) without dropping the dice. The state
        space includes the joint poses and velocities of the hand as well as the pose of the dice. The control space
        includes the target joint poses which is wrapped in a proportional controller.

        This example is used to compare the performance of EMPPI against MPPI with perfect knowledge and MPPI with
        imperfect model information. Here, the uncertainty resides in the finger joint proportional actuator gains of
        the Shadow hand. We initialize $p(\bar{\theta})$ using a uniform distribution (see Table~\ref{table:param})
        where we sample the actuator gains. Figure~\ref{fig:act_gain_results} illustrates the performance of our method
        against MPPI with perfect model parameters and MPPI with imperfect model parameters. Our method is shown to
        perform comparably to MPPI with perfect knowledge even when the physical parameters are still incorrect. It
        takes around $4$ seconds for the estimate of the actuator gains to converge to the true value. During the time,
        the controller is still able to provide a robust signal that utilizes the ensemble of physics simulators with
        the variations in the actuator gain to improve the overall performance of the task.


        \begin{figure*}[ht!]
            \centerline{
                \includegraphics[width=0.22\linewidth]{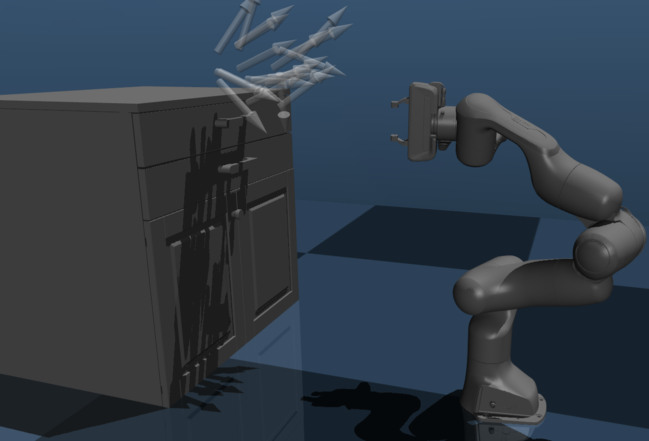}
                \includegraphics[width=0.22\linewidth]{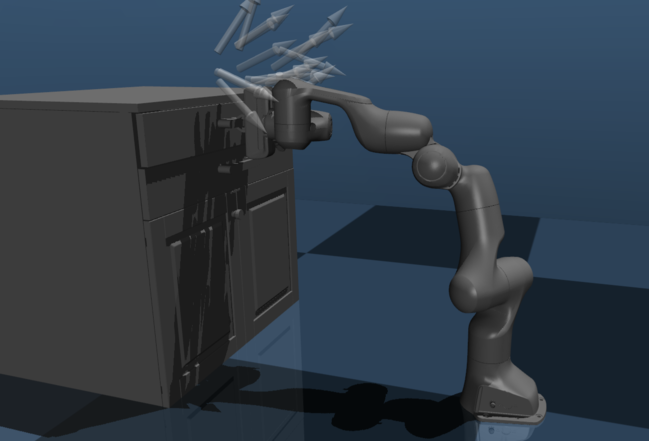}
                \includegraphics[width=0.22\linewidth]{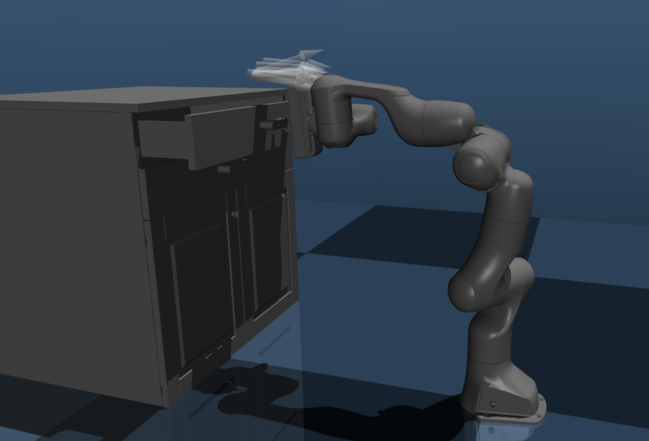}
                \includegraphics[width=0.22\linewidth]{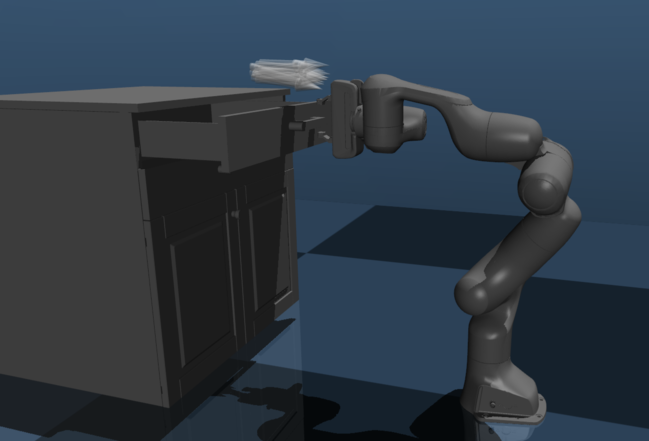}
            }
            \caption{
                Rendering of the model uncertainty in the cabinet articulation used for the Franka Panda robot arm
                example. Joints whose axis points perpendicular to the drawer are prismatic joints, all others are
                revolute joints. Our method synthesizes a control signal that generalizes to the uncertainty in the
                drawer's articulation. After detecting the movement of the handle, the distribution of parameters collapses
                to reflect the real-world cabinet.
            }
            \vspace{-5mm}
            \label{fig:franka_cabinet}
        \end{figure*}

        \textbf{Opening Doors with Adroit Hand} In this example, we deal with generalizing uncertainty of articulated
        objects in the environment. Specifically, we test our method against uncertainty in the articulation of doors
        and drawers. The task of the robot is to grab onto a handle and open the door when the robot does not know how
        the door will open. This is an important task as often robots do not have full knowledge of how doors are
        articulated in the world. We first test this in simulation using the Adroit hand door environment~\cite{6630771,
        Rajeswaran-RSS-18}. The state space includes the joint poses and velocities of the Adroit hand and
        arm as well as the hinge rotation (the location of the hinge is unknown). The control space includes the desired
        joint positions and the arm position in the world.

        Figure~\ref{fig:sim_results} illustrates a successful execution of the Adroit hand opening the door over a
        distribution of possible articulations depicted by the arrows. Each candidate joint in $p(\bar{\theta})$ has a
        joint axis and joint pose for the door where the value of $p(\theta_n)$ is illustrated using the transparency of
        the arrow. For the purpose of illustration we chose to sample each joint axis and pose from a uniform
        distribution over vertical and horizon poses, and a binomial distribution of axes spread away from the handle
        position. In Fig.~\ref{fig:sim_results}, we see that our method controls the Adroit hand toward the handle of
        the door, forcing an interaction with the door. This interaction provides state feedback through the actuator
        forces, the hand joint poses, and the handle position. The state feedback is then used to update the estimated
        door articulation as shown in the time series images in Fig.~\ref{fig:sim_results} where the hand is shown to
        successfully open the door.

    \subsection{Experiments}

        In this section, we provide three real-world experiments using EMPPI with the Franka Panda robot arm. The first
        experiment is planar pushing on a block with unknown mass and sliding friction towards a target location, second
        is object reconfiguration where the mass and the sliding friction of the object is unknown, and last is opening
        a cabinet drawer with uncertain articulation. Note that here, collisions are handled through a partial
        model\footnote{Compliance of the manipulation of objects not assumed and a partial model of the cabinet only
        consisting of a hand is used for handling collisions.} of the objects in the simulation environment. We refer
        the reader to the videos of each result in {\color{blue} \url{https://sites.google.com/view/emppi} },
        Table~\ref{table:param} for parameters and Fig.~\ref{fig:franka_cabinet} for the time series images of the
        experimental results.

        \textbf{Object Manipulation:} In this task, the goal is for the Panda robot to push an object towards a target
        where the mass and sliding friction of the object is unknown. The geometry of the object is loaded into the
        simulation where contacts are resolved. Therefore the state space includes the joint position and
        velocities of the robot and the position and orientation of the object. The object position and orientation are
        calculated using tracking tags. The generated control signal directly controls the joint
        velocities of the Panda robot. During the experimental trial, we found that the Panda arm would initially
        contact the object in a conservative manner, making light taps at the object to move it towards the target. As
        the state of the object and the robot are collected, the certainty in the sliding friction and the object mass
        continued enabling the robot arm to push the object towards the target location below the tag on the table. We
        found that there was significant coupling between the mass of the object and the sliding friction where the
        estimated mass was predicted to be $0.256$ kg where the actual mass was $0.164$ kg. This results in an apparent
        added mass due to the friction between the object and the table due to the coupling between mass and friction in
        rigid body movement. Regardless of the inaccuracies, having an ensemble of these models where the variations in
        the parameters is part of the control synthesis allows a robotic system to adapt to uncertain parameters and
        achieve the task in real-time. Furthermore, having physics simulations render the behavior of the object allows
        for a more structured approach to model-based control and is shown to be capable of online use without a loss of
        performance.

        \textbf{Object Reconfiguration:} In this example, the goal is for the Panda robot to rotate the object into a
        target configuration. The state and control spaces remain the same as with the prior experiment.
        As in the previous example, the uncertainty resides in the mass of the object and the friction between the
        object and the table. The goal is to rotate the object $90$ degrees on its side (chicken side up). This required
        EMPPI to generate actions for the Panda arm to compensate for uncertainty in the sliding of the object while not
        losing control of the object due to uncertainty in the mass. We found that the solution that EMPPI had come up
        with was to quickly brush across the top of the object and then tip the object on its side so that the object
        would naturally rotate over. This is an interesting solution for two reasons: the first is that EMPPI avoided
        actions which would unnecessarily slide the object, and the second is that the task had occurred quickly enough
        that the uncertainty in the parameters were unable to converge. Suggesting that EMPPI was able to plan for
        the uncertainty and generate actions which would have the best outcome despite the worse-case set of parameters
        which would cause an unsuccessful trial of the task. We recommend the reader see the results in the attached
        multimedia and in~{\color{blue} \url{https://sites.google.com/view/emppi} } .

        \textbf{Opening Drawer:} In this last example, the task is to open the drawer where there is uncertainty in how
        the drawer is articulated. The state space is the same as the prior experiment except the position
        and orientation of the handle is measured. The control space is converted to joint position control of the robot
        to ensure safe planning when interacting with the cabinet. We initialize the distribution of simulations using
        a normal distribution biased towards outward prismatic motion with a wide variance (see
        Table~\ref{table:param}). We found that this encourages pulling motions that do not damage the cabinet nor the
        robot while still providing a sufficient amount of samples. In Figure~\ref{fig:franka_cabinet}, we illustrate a
        depiction of the model simulations of the candidate joint axis. Note that axes whose largest unit value points
        upward or horizontally are specified as revolute joints where the unit values pointing outward perpendicular to
        the drawer are prismatic. In addition, the shown cabinet model is used as illustration, but does not exist in
        the models used for EMPPI. Instead, only the geometry of the handle is used for handling contacts where the
        position is handled through object tracking. During execution of the task, the uncertainty of the articulation
        does not change until the robot measures the movement of the drawer. At that time, the set of candidate
        parameters $\{\theta_n\}^N_{n=1}$ collapses into a distribution of prismatic joints (which reflects the
        real-world articulation).

\section{Conclusion} \label{sec:conclusion}

    In conclusion, we present a method that can synthesize control signals to complete a task while under parameter
    uncertainty. Our method utilizes the structure and complexity provided by physics engines to create a model-based
    controller that can generalize to parameter uncertainty. We show that our method is a natural extension to path
    integral control and is robust to various forms of uncertainty. We illustrate its effectiveness using a series of
    tasks with high dimensional robots. Last, our approach is experimentally validate our approach and provide some future
    work for improvements to the method.

\balance

\bibliographystyle{IEEEtran}

\bibliography{references}

\begin{thebibliography}{10}
\providecommand{\url}[1]{#1}
\csname url@rmstyle\endcsname
\providecommand{\newblock}{\relax}
\providecommand{\bibinfo}[2]{#2}
\providecommand\BIBentrySTDinterwordspacing{\spaceskip=0pt\relax}
\providecommand\BIBentryALTinterwordstretchfactor{4}
\providecommand\BIBentryALTinterwordspacing{\spaceskip=\fontdimen2\font plus
\BIBentryALTinterwordstretchfactor\fontdimen3\font minus
  \fontdimen4\font\relax}
\providecommand\BIBforeignlanguage[2]{{%
\expandafter\ifx\csname l@#1\endcsname\relax
\typeout{** WARNING: IEEEtran.bst: No hyphenation pattern has been}%
\typeout{** loaded for the language `#1'. Using the pattern for}%
\typeout{** the default language instead.}%
\else
\language=\csname l@#1\endcsname
\fi
#2}}

\bibitem{tan18rss}
J.~Tan, T.~Zhang, E.~Coumans, A.~Iscen, Y.~Bai, D.~Hafner, S.~Bohez, and
  V.~Vanhoucke, ``Sim-to-real: Learning agile locomotion for quadruped
  robots,'' in \emph{Proceedings of Robotics: Science and Systems}, June 2018.

\bibitem{peng2018sim}
X.~B. Peng, M.~Andrychowicz, W.~Zaremba, and P.~Abbeel, ``Sim-to-real transfer
  of robotic control with dynamics randomization,'' in \emph{IEEE International
  Conference on Robotics and Automation}, 2018, pp. 1--8.

\bibitem{jakobi1995noise}
N.~Jakobi, P.~Husbands, and I.~Harvey, ``Noise and the reality gap: The use of
  simulation in evolutionary robotics,'' in \emph{European Conference on
  Artificial Life}, 1995, pp. 704--720.

\bibitem{tobin2017iros}
J.~Tobin, R.~Fong, A.~Ray, J.~Schneider, W.~Zaremba, and P.~Abbeel, ``Domain
  randomization for transferring deep neural networks from simulation to the
  real world,'' in \emph{IEEE/RSJ International Conference on Intelligent
  Robots and Systems}, 2017, pp. 23--30.

\bibitem{chebotar2018closing}
Y.~Chebotar, A.~Handa, V.~Makoviychuk, M.~Macklin, J.~Issac, N.~Ratliff, and
  D.~Fox, ``Closing the sim-to-real loop: Adapting simulation randomization
  with real world experience,'' \emph{arXiv preprint arXiv:1810.05687}, 2018.

\bibitem{williams2017information}
G.~Williams, N.~Wagener, B.~Goldfain, P.~Drews, J.~M. Rehg, B.~Boots, and E.~A.
  Theodorou, ``Information theoretic {MPC} for model-based reinforcement
  learning,'' in \emph{IEEE International Conference on Robotics and
  Automation}, 2017.

\bibitem{williams2016aggressive}
G.~Williams, P.~Drews, B.~Goldfain, J.~M. Rehg, and E.~A. Theodorou,
  ``Aggressive driving with model predictive path integral control,'' in
  \emph{IEEE International Conference on Robotics and Automation}, 2016, pp.
  1433--1440.

\bibitem{williams2018robust}
G.~Williams, B.~Goldfain, P.~Drews, K.~Saigol, J.~Rehg, and E.~A. Theodorou,
  ``Robust sampling based model predictive control with sparse objective
  information,'' in \emph{Robotics Science and Systems}, 2018.

\bibitem{theodorou2010generalized}
E.~Theodorou, J.~Buchli, and S.~Schaal, ``A generalized path integral control
  approach to reinforcement learning,'' \emph{Journal of Machine Learning
  Research}, vol.~11, no. Nov, pp. 3137--3181, 2010.

\bibitem{schulman2015trust}
J.~Schulman, S.~Levine, P.~Abbeel, M.~Jordan, and P.~Moritz, ``Trust region
  policy optimization,'' in \emph{International Conference on Machine
  Learning}, 2015, pp. 1889--1897.

\bibitem{schulman2017proximal}
J.~Schulman, F.~Wolski, P.~Dhariwal, A.~Radford, and O.~Klimov, ``Proximal
  policy optimization algorithms,'' \emph{arXiv preprint arXiv:1707.06347},
  2017.

\bibitem{rajeswaran2017ICLR}
A.~Rajeswaran, S.~Ghotra, B.~Ravindran, and S.~Levine, ``Epopt: Learning robust
  neural network policies using model ensembles,'' in \emph{International
  Conference on Learning Representations (ICLR)}, 2016.

\bibitem{tassa2014control}
Y.~Tassa, N.~Mansard, and E.~Todorov, ``Control-limited differential dynamic
  programming,'' in \emph{IEEE International Conference on Robotics and
  Automation}, 2014, pp. 1168--1175.

\bibitem{li2004iterative}
W.~Li and E.~Todorov, ``Iterative linear quadratic regulator design for
  nonlinear biological movement systems.'' in \emph{International Conference on
  Informatics in Control, Automation and Robotics}, 2004, pp. 222--229.

\bibitem{tassa2012synthesis}
Y.~Tassa, T.~Erez, and E.~Todorov, ``Synthesis and stabilization of complex
  behaviors through online trajectory optimization,'' in \emph{IEEE/RSJ
  International Conference on Intelligent Robots and Systems}, 2012, pp.
  4906--4913.

\bibitem{stulp2011learning}
F.~Stulp, E.~Theodorou, J.~Buchli, and S.~Schaal, ``Learning to grasp under
  uncertainty,'' in \emph{IEEE International Conference on Robotics and
  Automation (ICRA)}, 2011, pp. 5703--5708.

\bibitem{becker2010motion}
A.~Becker and T.~Bretl, ``Motion planning under bounded uncertainty using
  ensemble control.'' in \emph{Robotics: Science and Systems}, 2010.

\bibitem{mordatch2015ensemble}
I.~Mordatch, K.~Lowrey, and E.~Todorov, ``Ensemble-cio: Full-body dynamic
  motion planning that transfers to physical humanoids,'' in \emph{IEEE/RSJ
  International Conference on Intelligent Robots and Systems (IROS)}, 2015, pp.
  5307--5314.

\bibitem{todorov2005generalized}
E.~Todorov and W.~Li, ``A generalized iterative {LQG} method for
  locally-optimal feedback control of constrained nonlinear stochastic
  systems,'' in \emph{Proceedings of the American Control Conference}, 2005,
  pp. 300--306.

\bibitem{Yu-RSS-17}
W.~Yu, J.~Tan, C.~K. Liu, and G.~Turk, ``Preparing for the unknown: Learning a
  universal policy with online system identification,'' in \emph{Proceedings of
  Robotics: Science and Systems}, 2017.

\bibitem{yu2018policy}
W.~Yu, C.~K. Liu, and G.~Turk, ``Policy transfer with strategy optimization,''
  in \emph{International Conference on Learning Representations}, 2019.

\bibitem{theodorou2012relative}
E.~A. Theodorou and E.~Todorov, ``Relative entropy and free energy dualities:
  Connections to path integral and kl control,'' in \emph{IEEE Conference on
  Decision and Control (CDC)}, 2012, pp. 1466--1473.

\bibitem{chebotar2017path}
Y.~Chebotar, M.~Kalakrishnan, A.~Yahya, A.~Li, S.~Schaal, and S.~Levine, ``Path
  integral guided policy search,'' in \emph{IEEE International Conference on
  Robotics and Automation}, 2017, pp. 3381--3388.

\bibitem{deisenroth2011pilco}
M.~Deisenroth and C.~E. Rasmussen, ``Pilco: A model-based and data-efficient
  approach to policy search,'' in \emph{Proceedings of the 28th International
  Conference on machine learning (ICML-11)}, 2011, pp. 465--472.

\bibitem{speekenbrink2016tutorial}
M.~Speekenbrink, ``A tutorial on particle filters,'' \emph{Journal of
  Mathematical Psychology}, vol.~73, pp. 140--152, 2016.

\bibitem{plappert2018multi}
M.~Plappert, M.~Andrychowicz, A.~Ray, B.~McGrew, B.~Baker, G.~Powell,
  J.~Schneider, J.~Tobin, M.~Chociej, P.~Welinder, \emph{et~al.}, ``Multi-goal
  reinforcement learning: Challenging robotics environments and request for
  research,'' \emph{arXiv preprint arXiv:1802.09464}, 2018.

\bibitem{6630771}
V.~{Kumar}, Z.~{Xu}, and E.~{Todorov}, ``Fast, strong and compliant pneumatic
  actuation for dexterous tendon-driven hands,'' in \emph{IEEE International
  Conference on Robotics and Automation}, 2013, pp. 1512--1519.

\bibitem{Rajeswaran-RSS-18}
A.~Rajeswaran, V.~Kumar, A.~Gupta, G.~Vezzani, J.~Schulman, E.~Todorov, and
  S.~Levine, ``Learning complex dexterous manipulation with deep reinforcement
  learning and demonstrations,'' in \emph{Proceedings of Robotics: Science and
  Systems}, 2018.

\end{thebibliography}

\end{document}